\documentclass[conference]{IEEEtran}
\ifCLASSINFOpdf
\else
\fi

\usepackage{algorithm,algorithmic}
\usepackage{color}
\usepackage{url}
\usepackage{times, epsfig, amsmath, theorem, amssymb,graphicx, wrapfig}

{\theorembodyfont{\rmfamily}

}

\newtheorem{definition}{Definition}

\newtheorem{observation}{Observation}

{\theorembodyfont{\rmfamily}
\newtheorem{example}{Example}
}

\newcommand{\prob}{\operatorname{prob}}
\newcommand{\MH}{\operatorname{PAMH}}

\renewcommand{\emph}{\textit}
\renewcommand{\em}{\textit}

\begin{document}
%
\title{Causal Decision Trees}

\author{\IEEEauthorblockN{Jiuyong Li, Saisai Ma, Thuc Duy Le, Lin Liu and Jixue Liu}
\IEEEauthorblockA{School of Information Technology and Mathematical Sciences, University of South Australia, Australia\\
Mawson Lakes, SA 5095, Australia}} 


%


\maketitle

\begin{abstract}
Uncovering causal relationships in data is a major objective of data analytics. Causal relationships are normally discovered with designed experiments, e.g. randomised controlled trials, which, however are expensive or infeasible to be conducted in many cases. Causal relationships can also be found using some well designed observational studies, but they require domain experts' knowledge and the process is normally time consuming. Hence there is a need for scalable and automated methods for causal relationship exploration in data. Classification methods are fast and they could be practical substitutes for finding causal signals in data. However, classification methods are not designed for causal discovery and a classification method may find false causal signals and miss the true ones. In this paper, we develop a causal decision tree where nodes have causal interpretations. Our method follows a well established causal inference framework and makes use of a classic statistical test. The method is practical for finding causal signals in large data sets.
\end{abstract}


%

\section{Introduction}
Detecting causal relationships in data is an important data analytics task as causal relationships can provide better insights into data, as well as actionable knowledge for correct decision making and timely intervening in processes at risk.

Causal relationships are normally identified with experiments, such as randomised controlled trials~\cite{Shadish2002}, which are effective but expensive and often impossible to be conducted. Causal relationships can also be found by observational studies, such as cohort studies and case control studies~\cite{Rosenbaum2010}. An observational study takes a causal hypothesis and tests it using samples selected from historical data or collected passively over the period of time when observing the subjects of interest. Therefore observational studies need domain experts' knowledge and interactions in data selection or collection and the process is normally time consuming.

Currently there is a lack of scalable and automated methods for causal relationship exploration in data. These methods should be able to find causal signals in data without requiring domain knowledge or any hypothesis established beforehand. The methods must also be efficient to deal with the increasing amount of big data.

Classification methods are fast and have the potential to become practical substitutes for finding causal signals in data since the discovery of causal relationships is a type of supervised learning when the target or outcome variable is fixed. Decision trees~\cite{DataMiningBookHan2011} are a good example of classification methods, and they have been widely used in many areas, including social and medical data analyses.

\begin{wrapfigure}{r}{0.25\textwidth}
\begin{center}
      \includegraphics[width=0.25\textwidth]{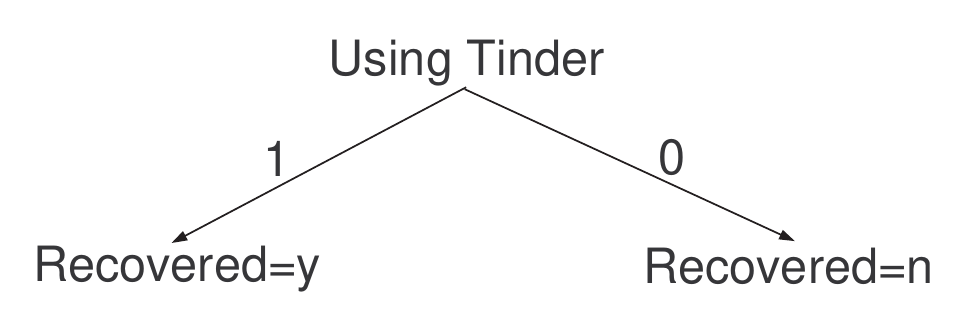}
\end{center}
\caption{A simple decision tree}
\label{fig_hypnotizedDT}
\end{wrapfigure}

However, classification methods are not designed with causal discovery in mind and a classification method may find false causal signals in data and miss true causal signals. For example, Figure~\ref{fig_hypnotizedDT} shows a decision tree built from a hypothesised data set of the recovery of a disease. Based on the decision tree, we may conclude that the use of Tinder (a matchmaking mobile app) helps cure the disease. However, it is misleading since the majority of people using Tinder are young whereas most people not using Tinder are old. Young people will recover from the disease anyway and old people have a lower chance of recovery. This misleading decision tree is caused by an unfair comparison between the two different groups of people. It may be a good classification tree to predict the likelihood of recovery, but it does not imply the causes of recovery and its nodes do not have any causal interpretation.

A classification method fails to take account of the effects of other variables on the  class or outcome variable when examining the relationship between a variable and the class variable, and this is the major reason for the false discoveries (of causal relationships). For example, when we study the relationship between using Tinder and the recovery of a disease, the effect of other variables such as age, gender, and health condition of patients (who may or may not use Tinder) should be considered. The objective is not simply to maximise the difference of the conditional probabilities of recovered and not recovered conditioning on the use of Tinder when a classifier is being sought.

In this paper, we design a causal decision tree (CDT) where nodes have causal interpretations. As presented in the following sections, our method follows a well established causal inference framework, the potential outcome model, and it makes use of a classic statistical test, Mantel-Haenszel test. The proposed CDT is practical for uncovering causal signals in large data.

The paths in a CDT are not interpreted as ``if - then'' first order logic rules as in a normal decision tree. For example, Figure~\ref{fig_Titanic_Adult} (left) shows a CDT learnt from the Titanic data set. It does not read as ``if female = n then survived = n''. A node in a CDT indicates a causal factor of the outcome attribute. The node `female' indicates that being female or not is causally related to survived or not; the node `thirdClass' shows that in the female group (the context), staying in a third class cabin or not is causally related to survived or not.

The main contributions of this paper are as follows:

\begin{itemize}
\item We systematically analyse the limitations of decision trees for causal discovery and identify the underlying reasons.
\item We propose the CDT method, which can be used to represent and identify simple and interpretable causal relationships in data, including big data.
\end{itemize}

\section{Cause and effect in the potential outcome framework}
\label{sec_potentialoutcome}

Let $X$ be a predictive attribute and $Y$ the outcome attribute where $x \in \{0, 1\}$ and $y \in \mathbb{R}_{\geq 0}$
We aim to find out if there is a causal relationship between $X$ and $Y$. For easy discussion, we consider that $X=1$ is a treatment and $Y=1$ the recovery. We will establish if the treatment is effective for the recovery.

The potential outcome or counterfactual model \cite{Pearl2000,Morgan2007} is a well established framework for causal inference. Here we introduce the basic concepts of the model and a principle for estimating the average causal effect, mainly following the introduction in~\cite{morgan2006matching}.

With the potential outcome model, an individual $i$ in a population has two potential outcomes for a treatment $X$:  $Y^1_i$ when taking the treatment and $Y^0_i$ when not taking the treatment. We say that $Y^1_i$ is the potential outcome in the treatment state and $Y^0_i$ is the potential outcome in the control state. Then we have the following definition.
\begin{definition}[Individual level causal effect (ICE)]
The individual level causal effect is defined as the difference of two potential outcomes of an individual, i.e. $\delta_i = Y^1_i - Y^0_i$.
\end{definition}
In practice we can only find out one outcome $Y^1_i$ or $Y^0_i$ since one person can be placed in either the treatment group ($X=1$) or the control group ($X=0$). One of the two potential outcomes has to be estimated. So the potential outcome model is also called counterfactual model. For example, we know that Mary has a headache (the outcome) and she did not take aspirin (the treatment), i.e. we know $Y^0_i$. The question is what the outcome would be if Mary took aspirin one hour ago, i.e. we want to know $Y^1_i$ and to estimate the ICE of aspirin on Mary's condition (having headache or not).

If we had both $Y^1_i$ and $Y^0_i$ of an individual we would aggregate the causal effects of individuals in a population to get the average causal effect as defined below, where $E[.]$ stands for expectation operator in probability theory.

\vspace{-0.1cm}
\begin{definition}[Average causal effect (ACE)]
The average causal effect of a population is the average of the individual level causal effects in the population, i.e. $E[\delta_i] = E[Y^1_i] - E[Y^0_i]$.
\end{definition}

\vspace{-0.1cm}
Note that $i$ is kept in the above formula as other work in the counterfactual framework to indicate individual level heterogeneity of potential outcomes and causal effects.

Assuming that $\pi$ proportion of samples take the treatment and $(1-\pi)$ proportion do not, and the sample size is large so the error caused by sampling is negligible, given a data set $\textbf{D}$, the ACE, $E[\delta_i]$ can be estimated as:
\setlength{\belowdisplayskip}{3pt} \setlength{\belowdisplayshortskip}{3pt}
\setlength{\abovedisplayskip}{3pt} \setlength{\abovedisplayshortskip}{3pt}
\begin{eqnarray}
\label{eqn_1}
E_\textbf{D}[\delta_i] & \!\!\!= \!\!\!& \pi (E_\textbf{D}[Y^1_i | X_i = 1] - E_\textbf{D}[Y^0_i | X_i = 1]) + \nonumber \\
 &  & (1-\pi) (E_\textbf{D}[Y^1_i | X_i = 0] - E_\textbf{D}[Y^0_i | X_i = 0])
\end{eqnarray}
That is, the ACE of the population is the ACE in the treatment group plus the ACE in the control group, where $X_i = 1$ indicates that an individual takes the treatment and the causal effect is $(Y^1_i | X_i = 1) - (Y^0_i | X_i = 1)$. Similarly, $X_i = 0$ indicates that an individual does not take the treatment and the causal effect is $(Y^1_i | X_i = 0) - (Y^0_i | X_i = 0)$.

In a data set, we can observe the potential outcomes in the treatment state for those treated, ($Y^1_i | X_i = 1$), and the potential outcomes in the control state for those not treated, ($Y^0_i | X_i = 0$). However, we cannot observe the potential outcomes in the control state for those treated, ($Y^0_i | X_i = 1$), or the potential outcomes in the treatment state for those not treated, ($Y^1_i | X_i = 0$). We have to estimate what the potential outcome, $(Y^0_i | X_i = 1)$, would be if an individual did not take the treatment (in fact she has); and what potential outcome, ($Y^1_i | X_i = 0$), would be if an individual took the treatment (in fact she has not).

With a data set $\textbf{D}$ we can obtain the following ``na{\"i}ve'' estimation of the ACE:
\begin{eqnarray}
\label{eqn_2}
E^{naive}_\textbf{D}[\delta_i] = E_\textbf{D}[Y^1_i | X_i = 1] - E_\textbf{D}[Y^0_i | X_i = 0]
\end{eqnarray}

The question is when the na{\"i}ve estimation~(Equation (\ref{eqn_2})) will approach the true estimation~(Equation (\ref{eqn_1})).

If the assignment of individuals to the treatment and control groups is purely random, the estimation in Equation~(2) approaches the estimation in Equation~(1).
In an observational data set, however, the random assignment is not possible. How can we estimate the average causal effect? A solution is by perfect stratification. Let the differences of individuals in a data set be characterised by a set of attributes $\textbf{S}$ (excluding $X$ and $Y$) and let the data set be perfectly stratified by $\textbf{S}$. In each stratum, apart from the fact of taking treatment or not, all individuals are indistinguishable from each other. Under the perfect stratification assumption, we have:
\begin{eqnarray}
\label{eqn_3}
E[Y^1_i | X_i = 0, \textbf{S}=\textbf{s}_i] = E[Y^1_i | X_i = 1, \textbf{S} = \textbf{s}_i]   \label{firsteqn}\\
E[Y^0_i | X_i = 1, \textbf{S} = \textbf{s}_i] = E[Y^0_i | X_i = 0, \textbf{S}=\textbf{s}_i]\label{secondeqn}
\end{eqnarray}

\noindent where $\textbf{S}=\textbf{s}_i$ indicates a stratum of perfect stratification. Since individuals are indistinguishable in the stratum, unobserved potential outcomes can be estimated by observed ones. Specifically, the mean potential outcome in the treatment state for those untreated is the same as that in the treatment state for those treated (Equation~ (\ref{firsteqn})), and the mean potential outcome in the control state for those treated is the same as that in the control state for those untreated (Equation~(\ref{secondeqn})). By replacing Equation~(1) with Equations~(\ref{firsteqn}) and (\ref{secondeqn}), we have:
\begin{eqnarray}
 & & E_\textbf{D}[\delta_i | \textbf{S} = \textbf{s}_i] \nonumber\\
 & = & \pi (E_\textbf{D}[Y^1_i | X_i = 1, \textbf{S} = \textbf{s}_i] - E_\textbf{D}[Y^0_i | X_i = 1, \textbf{S} = \textbf{s}_i]) +  \nonumber \\
 &   & (1-\pi) (E_\textbf{D}[Y^1_i | X_i = 0, \textbf{S} = \textbf{s}_i] - E_\textbf{D}[Y^0_i | X_i = 0, \textbf{S} = \textbf{s}_i])  \nonumber \\
 & = & \pi (E_\textbf{D}[Y^1_i | X_i = 1, \textbf{S} = \textbf{s}_i] - E_\textbf{D}[Y^0_i | X_i = 0, \textbf{S} = \textbf{s}_i]) +  \nonumber \\
 &   & (1-\pi) (E_\textbf{D}[Y^1_i | X_i = 1, \textbf{S} = \textbf{s}_i] - E_\textbf{D}[Y^0_i | X_i = 0, \textbf{S} = \textbf{s}_i])  \nonumber \\
 & = & E_\textbf{D}[Y^1_i | X_i = 1, \textbf{S} = \textbf{s}_i] - E_\textbf{D}[Y^0_i | X_i = 0, \textbf{S} = \textbf{s}_i]  \nonumber \\
 & = & E^{naive}_\textbf{D}[\delta_i | \textbf{S} = \textbf{s}_i] \nonumber
\end{eqnarray}

As a result, the na{\"i}ve estimation approximates the true average causal effect, and we have the following observation.

\vspace{-0.25cm}
\begin{observation}\label{principle}[Principle for estimating average causal effect] The average causal effect can be estimated by taking weighted sum of na{\"i}ve estimators in stratified sub data sets.
\end{observation}

\vspace{-0.1cm}
This principle ensures that each comparison is between individuals with no observable differences, and hence the estimated causal effect is not resulted from other factors than the studied one. In the following, we will use this principle to estimate causal effect in observational data sets.

\section{A decision tree may not encode \\ causal relationships}
Decision trees are a popular classification model, with two types of nodes: branching and leaf nodes. A branching node represents a predictive attribute and each of its values denotes a choice and leads to another branching node or a leaf node representing a class. Now we use the potential outcome model to explain why decision trees may not encode causality.
\begin{figure}[t]
\center
\begin{tabular}{cc}
\begin{minipage}[htbp]{0.25 \textwidth}
\begin{tabular}{|cccccc|}
\hline
\scriptsize A & \scriptsize B & \scriptsize C & ~ & \scriptsize Y & \scriptsize count\\[-1pt]
\hline
\scriptsize 0 & \scriptsize 1 & \scriptsize 0 & ~ & \scriptsize 0 & \scriptsize 20 \\[-2pt]
\scriptsize 0 & \scriptsize 0 & \scriptsize 1 & ~ & \scriptsize 1 & \scriptsize 20 \\[-2pt]
\scriptsize 1 & \scriptsize 1 & \scriptsize 1 & ~ & \scriptsize 1 & \scriptsize 10\\[-2pt]
\scriptsize 1 & \scriptsize 0 & \scriptsize 1 & ~ & \scriptsize 1 & \scriptsize 10 \\[-2pt]
\scriptsize 1 & \scriptsize 0 & \scriptsize 0 & ~ & \scriptsize 1 & \scriptsize 20 \\
\hline
\end{tabular}
\end{minipage}
&
\begin{minipage}[htbp]{0.11\textwidth}
      \includegraphics[width=\textwidth]{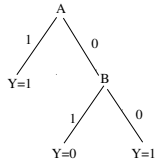}
\end{minipage}
\\
~ & ~ \\
(a) & (b) \\
\end{tabular}
\vspace{-0.15cm}
\caption{An example showing that a decision tree does not encode causal relationships (a) An exemplar data set. (b) A decision tree of the data set}
\vspace{-0.55cm}
\label{fig_ex-decisionTree1}
\end{figure}

\vspace{-0.3cm}
\begin{example}
\label{ex_non-causaltree}
Given the data set and its corresponding decision tree as in Figure~\ref{fig_ex-decisionTree1}. The decision tree has perfectly explained the data set, but the tree does not represent causal relationships.

Path $A=1 \to Y=1$, with the difference in probabilities, $(\prob(Y=1 | A=1) - \prob(Y=0 | A=1))= 1$, represents a top quality discriminative rule. 
Let us assume that $A=1$ is a treatment and $Y$ represents the outcome. To derive the average causal effect of $A$ on $Y$, we need to stratify the data set such that in each stratum the records are exchangeable. Here $\{B, C\}$ are the stratifying attributes. The data set in Figure~\ref{fig_ex-decisionTree1} is stratified into four strata and their summaries are as follows:

\vspace{0.1cm}
\begin{center}
\noindent \begin{tabular}{|c|cc|}
\hline
\scriptsize $\{B, C\}$ & \multicolumn{2}{|c|}{\scriptsize  Y} \\[-2pt]
 \scriptsize $\{0, 0\}$ &\scriptsize  $1$ &\scriptsize  $0$ \\[-2pt]
\hline
\scriptsize $A=1$ & \scriptsize 20 &\scriptsize  0 \\[-2pt]
\scriptsize $A=0$ & \scriptsize 0 &\scriptsize  0 \\
\hline
\end{tabular}
\hspace*{0.05 cm}
\begin{tabular}{|c|cc|}
\hline
\scriptsize $\{B, C\}$ & \multicolumn{2}{|c|}{\scriptsize Y} \\[-2pt]
\scriptsize $\{0, 1\}$ &\scriptsize  $1$ &\scriptsize  $0$ \\[-2pt]
\hline
\scriptsize $A=1$ &\scriptsize  10 &\scriptsize  0 \\[-2pt]
\scriptsize $A=0$ & \scriptsize 20 &\scriptsize  0 \\
\hline
\end{tabular}
\\
\begin{tabular}{|c|cc|}
\hline
\scriptsize $\{B, C\}$ & \multicolumn{2}{|c|}{\scriptsize Y} \\[-2pt]
\scriptsize $\{1, 0\}$ & \scriptsize $1$ &\scriptsize  $0$ \\[-2pt]
\hline
\scriptsize $A=1$ &\scriptsize  0 &\scriptsize  0 \\[-2pt]
\scriptsize $A=0$ &\scriptsize  0 & \scriptsize 20 \\
\hline
\end{tabular}
\hspace*{0.05 cm}
\begin{tabular}{|c|cc|}
\hline
\scriptsize $\{B, C\}$ & \multicolumn{2}{|c|}{\scriptsize Y} \\[-2pt]
\scriptsize $\{1, 1\}$ & \scriptsize $1$ & \scriptsize $0$ \\[-2pt]
\hline
\scriptsize $A=1$ & \scriptsize 10 &\scriptsize  0 \\[-2pt]
\scriptsize $A=0$ &\scriptsize  0 &\scriptsize  0 \\
\hline
\end{tabular}
\end{center}
\normalsize

\vspace{0.1cm}
In the first stratum, all records have $B=0$ and $C=0$. There are no records from the control group ($A=0$), hence we cannot estimate the average causal effect in this stratum. Similarly, we cannot estimate causal effects from the third ($B=1$, $C=0$) and fourth ($B=1$, $C=1$) strata. In the second stratum ($B=0$, $C=1$), $E_\textbf{D}[Y^1 | A = 1] - E_\textbf{D}[Y^0 | A = 0] = 1-1 =0$. All cases regardless they are treated or not treated have the same outcome. So the causal relationship between $A$ and $Z$ cannot be established.

For paths $(A=0, B=1) \to Y=0$ and $(A=0, B=0) \to Y=1$, let us try to establish a causal relationship between $B$ and $Y$ in the sub data set where $A=0$.

The two strata by attribute $C$ are summarised as:

\vspace{0.1cm}
\begin{center}
\begin{tabular}{|c|cc|}
\hline
\scriptsize $C$ & \multicolumn{2}{|c|}{\scriptsize Y} \\[-3pt]
 \scriptsize $0$ & \scriptsize $1$ & \scriptsize $0$ \\[-2pt]
\hline
\scriptsize $B=1$ & \scriptsize 0 &\scriptsize  20 \\[-2pt]
\scriptsize $B=0$ &\scriptsize  0 & \scriptsize 0 \\
\hline
\end{tabular}
\hspace*{0.2 cm}
\begin{tabular}{|c|cc|}
\hline
\scriptsize $C$ & \multicolumn{2}{|c|}{\scriptsize Y} \\[-3pt]
\scriptsize $1$ & \scriptsize $1$ & \scriptsize $0$ \\[-2pt]
\hline
\scriptsize $B=1$ &\scriptsize  0 &\scriptsize  0 \\[-2pt]
\scriptsize $B=0$ &\scriptsize  20 & \scriptsize 0 \\
\hline
\end{tabular}
\end{center}

\vspace{0.1cm}
In the two strata ($C\!=\!0$ and $C\!=\!1$) there are only cases in either the treatment group ($B\!\!=\!\!1$) or the control group ($B\!=\!0$). There is no way to estimate the average causal effect, so we cannot establish a causal relationship between $B$ and $Y$.

In this example, we see that a perfect decision tree does not indicate any causal relationship. In other words, in this data set, there is not enough evidence to support causal relationships.
\end{example}

\section{From normal decision trees to causal decision trees}
\label{sec_causalDecisionTree}

Let $\textbf{X} = \{X_1, X_2, \ldots, X_m\}$ be a set of predictive attributes where $x_i \in \{0, 1\}$ for $1 \le i \le m$, and $Y$ be an outcome attribute where $y \in \{0, 1\}$. Data set $\textbf{D}$ contains $n$ records taking various assignments of values for $\textbf{X}$ and $Y$, each of which represents the record of an observation. Let us assume that $\textbf{X}$ includes all the attributes for characterising an individual, and the data set is large and thus there is no bias in the sampling process.

\subsection{Why a decision tree may not encode causal relationships?}
The construction of a decision tree follows a divide and conquer strategy, and the most important decision to be made in the construction is to choose which attribute as a branching node.
Information gain, information gain ratio or Gini index can be used to choose a branching node~\cite{DataMiningBookHan2011}. These criteria have slight differences, but they all aimed at finding a discriminative attribute in the context.

\vspace{-0.32cm}
\begin{definition}[Discriminative attribute]
Given a data set $\textbf{D}'$, a discriminative attribute is the attribute $X_i$ such that $|\prob(Y=1|X_i$ $ = 1)-\prob(Y=0|X_i = 1)|$ is maximised.
\end{definition}

Note that $\textbf{D}'$ is a sub data set (of $\textbf{D}$) defined by the attribute values in the prefix path of the current branching node under consideration. It is a context specific data set (see Section~\ref{sec_causalDT} for details).

In the following, we will discuss why a decision tree may not represent causal relationships.

Firstly, The objective of a discriminative attribute (maximising $\prob(Y=1|X_i = 1) - \prob(Y=0|X_i = 1)$) is different from that of a causal factor (having significant causal effect  $\prob(Y=1|X_i = 1) - \prob(Y=1|X_i = 0)$).

Secondly, the estimation of $\prob(Y\!=\!1|X_i\!=\!1)\!-\!\prob(Y\!=\!0|X_i\!=\!1)$ for choosing a discriminative attribute is based on the data set $\textbf{D}'$, while the estimation of the causal effect of $X_i$ on $Y$ is based on the stratified data set $\textbf{D}_{\textbf{S} = \textbf{s}_i}$ to avoid unfair comparison. For example, let $X_i$ be a treatment and $Y$ the recovery, the comparison between individuals with and without treatments have to be in the same age and gender groups and have similar medical conditions. Otherwise, the comparison is meaningless. In other words, when a comparison is within a stratified data set, the effect of other attributes on $Y$ is eliminated and hence the difference ($\prob(Y\!=\!1|X_i\!=\!1)\!-\!\prob(Y\!=\!0|X_i\!=\!1)$)  reflects the causal effect of $X_i$ on $Y$.

Essentially the main limitation of a decision tree is that it does not consider other attributes in determining a branching attribute. The choice of a branching attribute does not rely on the causal effect of the attribute on the outcome attribute.

Following Observation~\ref{principle}, the estimation of causal effect should be based on stratified data where the difference of individuals in a stratum is eliminated. 

\vspace{-0.1cm}
\subsection{A measure for causal effect}
\label{sec_stratificationandmeasure}

Based on the previous discussion, to estimate the causal effect of a predictive attribute $Q$ on the outcome $Y$, we stratify a data set using  $\textbf{X} \backslash \{Q\}$ so that within each stratum there is no observable difference among the records.

A measure of average causal effect should be able to quantify the difference of outcomes in two groups (treatment and control). For binary outcomes, odds ratio \cite{Fleiss2003} is suitable for measuring the difference of two outcomes. Let the following table summarise the statistics of stratum $\textbf{s}_k$ where $\textbf{S} = \textbf{s}_k$.

\vspace{0.05cm}
\begin{center}
\begin{tabular}{|c|cc|c|}
\hline
\scriptsize $\textbf{s}_k$ &\scriptsize  $Y=1$ &\scriptsize  $Y=0$ &\scriptsize  total \\[-2pt]
\hline
\scriptsize $Q=1$ &\scriptsize  $n_{11k}$ & \scriptsize $n_{12k}$ &\scriptsize  $n_{1k}$ \\[-2pt]
\scriptsize $Q=0$ & \scriptsize $n_{21k}$ &\scriptsize  $n_{22k}$ &\scriptsize  $n_{2k}$ \\[-1pt]
\hline
\scriptsize total &\scriptsize  $n_{.1k}$ & \scriptsize $n_{.2k}$ &\scriptsize  $n_{..k}$ \\
\hline
\end{tabular}
\end{center}

\vspace{0.1cm}
The odds ratio (measuring the difference of $Y$ between groups $Q=1$ and $Q=0$) of the stratum $k$ is $(n_{11k}n_{22k}) / (n_{12k}n_{21k})$ or equivalently, $\ln(n_{11k})+\ln(n_{22k})-\ln(n_{12k})- \ln(n_{21k})$.

A question is how to get the aggregated difference over all the strata of a data set. Partial association test~\cite{Birch:Partial} is a means to achieve this. Over all the $r$ strata of a data set, the difference can be summarised as:
\begin{eqnarray} \label{MHstatistic}
 \MH (Q, Y) =  \frac{(|\sum_{k=1}^r \frac{n_{11k}n_{22k}-n_{21k}n_{12k}}{n_{..k}}|-\frac{1}{2})^2} {\sum_{k=1}^r \frac{n_{1.k}n_{2.k}n_{.1k}n_{.2k}}{n_{..k}^2(n_{..k} -1)}}
\end{eqnarray}

This is the test statistic of the Mantel-Haenszel test \cite{MHtest1959,Birch:Partial}. The purpose of the test is to see if the association between $Q$ and $Y$ is consistent in all conditions (strata). Such consistency is a strong indication of direct causal relationship between two variables~\cite{Birch:Partial}. The  test statistic has a Chi-square distribution (degree of freedom=1). Given a significance level $\alpha$, if $\MH(Q, Y) \ge \chi^2_{\alpha}$, the null hypothesis that $Q$ and $Y$ are independent in all strata is rejected and the partial association between $Q$ and $Y$ is significant. An example of Mantel-Haenszel test is given in Example~\ref{ex_causaltree} in the next section.

\subsection{Causal decision trees}
\label{sec_causalDT}

Our aim is to build a causal decision tree (CDT) where a non-leaf node represents a causal attribute, an edge denotes an assignment of a value of a causal attribute, and a leaf represents an assignment of a value of the outcome.  A path from the root to a leaf represents a series of assignments of values of the attributes and a highly probable outcome value as the leaf.

A CDT differs from a normal decision tree in that each of its non-leaf nodes has a causal interpretation with respect to the outcome, i.e. a non-leaf node and the outcome attribute have a context specific causal relationship as defined below.

\vspace{-0.3cm}
\begin{definition}[Context]
Let $\textbf{P} \subset \textbf{X}$, then a value assignment of $\textbf{P}$, $\textbf{P} = \textbf{p}$, is called a context and $(\textbf {D} | \textbf{P} = \textbf{p})$ is a context specific data set where $\textbf{P} = \textbf{p}$ holds for all records in $\textbf{D}$.
\end{definition}

\vspace{-0.4cm}
\begin{definition}[Context specific causal relationship]
Let $\textbf{P} = \textbf{p}$ be a context and $Q$ be a predictive attribute and $\textbf{P} \cap \{Q\} = \emptyset$. $Q$ and the outcome attribute $Y$ have a context specific causal relationship if $\MH(Q, Y)$ is greater than a threshold in the context specific data set $(\textbf {D} | \textbf{P} = \textbf{p})$.
\end{definition}

\vspace{-0.05cm}
A context specific causal relationship between the root node and the outcome attribute of a CDT is global or context free, i.e. the context attribute set $\textbf{P}$ is empty, and a context specific causal relationships between a non-root node $A$ and the outcome $Y$ is a refinement of the causal relationship between $A$'s parent and $Y$. For example, with the CDT in Figure~\ref{fig_Titanic_Adult} (right), the causal relationship between the root `age$<$30' and the outcome `$>$50K' is context free, while the causal relationship `education-num$>$12' having with the outcome is in the context of `age$<$30' being no, which is a refinement of the causal relationship `age$<$30' (parent of `education-num$>$12') having with the outcome and is a more specific but stronger relationship. 

\vspace{-0.3cm}
\begin{definition}[Causal decision tree (CDT)]
In a causal decision tree, a non-leaf node $Q$ represents a context specific causal relationship between $Q$ and the outcome attribute $Y$ where the context is a series of value assignments of the attributes along the path from the root and to the parent of $Q$. A leaf node represents a value assignment of $Y$, which is the most probable value of $Y$ in the context specific data set where the context is a series of value assignments of the attributes along the path from the root to the leaf.
\end{definition}

We use the following example to show that a CDT encodes causal relationships.

\begin{figure}
\center
\begin{tabular}{ccc}
\begin{minipage}[htbp]{0.25\textwidth}
\begin{tabular}{|cccccc|}
\hline
\scriptsize A &\scriptsize B &\scriptsize C & ~ &\scriptsize Y\scriptsize & \scriptsize count \\[-1pt]
\hline
\scriptsize 0 &\scriptsize  0 &\scriptsize  0 & ~ &\scriptsize  1 &\scriptsize  10 \\[-2pt]
\scriptsize 0 &\scriptsize  0 &\scriptsize  1 & ~ &\scriptsize  1 &\scriptsize  20 \\[-2pt]
\scriptsize 0 &\scriptsize  1 &\scriptsize  0 & ~ &\scriptsize  0 &\scriptsize  10 \\[-2pt]
\scriptsize 0 &\scriptsize  1 &\scriptsize  1 & ~ &\scriptsize  0 &\scriptsize  20 \\[-2pt]
\scriptsize 1 &\scriptsize  0 & \scriptsize 0 & ~ &\scriptsize  1 &\scriptsize  10 \\[-2pt]
\scriptsize 1 &\scriptsize  0 &\scriptsize  1 & ~ &\scriptsize  0 &\scriptsize  5 \\[-2pt]
\scriptsize 1 &\scriptsize  1 &\scriptsize  0 & ~ &\scriptsize  1 &\scriptsize  10 \\[-2pt]
\scriptsize 1 &\scriptsize  1 & \scriptsize 1 & ~ &\scriptsize  1 &\scriptsize  15 \\
\hline
\end{tabular}
\end{minipage}
&
\begin{minipage}[htbp]{0.12\textwidth}
      \includegraphics[width=\textwidth]{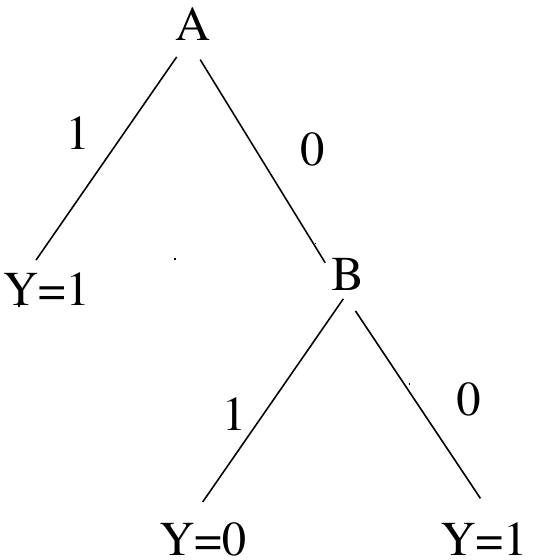}
\end{minipage}
\\
~ & ~ \\
(a) & (b) \\
\end{tabular}
\vspace{-0.1cm}
\caption{An example shows that a CDT represents causal relationships (a) An exemplar data set. (b) a CDT of the data}
\vspace{-0.5cm}
\label{fig_ex-decisionTree2}
\end{figure}

\vspace{-0.3cm}
\begin{example}
\label{ex_causaltree}

From the data set shown in Figure~\ref{fig_ex-decisionTree2}, for path $A=1 \to Y=1$ of the CDT, we have the following summaries of the strata in terms of attributes $\{B, C\}$:
\vspace{0.3cm}
\begin{center}
\begin{tabular}{|c|cc|}
\hline
\scriptsize $\{B, C\}$ & \multicolumn{2}{|c|}{\scriptsize Y} \\[-2pt]
 \scriptsize $\{0, 0\}$ &\scriptsize  $1$ &\scriptsize  $0$ \\[-2pt]
\hline
\scriptsize $A=1$ &\scriptsize  10 &\scriptsize  0 \\[-2pt]
\scriptsize $A=0$ &\scriptsize  10 &\scriptsize  0 \\
\hline
\end{tabular}
\hspace*{0.05 cm}
\begin{tabular}{|c|cc|}
\hline
\scriptsize $\{B, C\}$ & \multicolumn{2}{|c|}{\scriptsize Y} \\[-2pt]
\scriptsize $\{0, 1\}$ & \scriptsize $1$ &\scriptsize  $0$ \\[-2pt]
\hline
\scriptsize $A=1$ & \scriptsize 0 &\scriptsize  5 \\[-2pt]
\scriptsize $A=0$ &\scriptsize  20 &\scriptsize  0 \\
\hline
\end{tabular}
\\
\begin{tabular}{|c|cc|}
\hline
\scriptsize $\{B, C\}$ & \multicolumn{2}{|c|}{\scriptsize Y} \\[-2pt]
\scriptsize $\{1, 0\}$ & \scriptsize $1$ &\scriptsize  $0$ \\[-2pt]
\hline
\scriptsize $A=1$ &\scriptsize  10 &\scriptsize  0 \\[-2pt]
\scriptsize $A=0$ &\scriptsize  0 &\scriptsize  10 \\
\hline
\end{tabular}
\hspace*{0.05 cm}
\begin{tabular}{|c|cc|}
\hline
\scriptsize $\{B, C\}$ & \multicolumn{2}{|c|}{\scriptsize Y} \\[-2pt]
\scriptsize $\{1, 1\}$ & \scriptsize $1$ &\scriptsize  $0$ \\[-2pt]
\hline
\scriptsize $A=1$ &\scriptsize  15 &\scriptsize  0 \\[-2pt]
\scriptsize $A=0$ &\scriptsize  0 &\scriptsize  20 \\
\hline
\end{tabular}
\end{center}

\vspace{0.1cm}
We now calculate the Mantel-Haenszel test statistic (Equation (\ref{MHstatistic})),  $\MH(A, Y)$. The first table above (for the stratum $B=0$, $C=0$) does not contribute to the calculation of $\MH(A, Y)$ since it has one column of zero values.

In the stratum $B=0$ and $C=1$,
\[
\frac{n_{11k}n_{22k}-n_{21k}n_{12k}}{n_{..k}} = \frac{0*0-20*5}{25} = -4
\]
\[
\frac{n_{1.k}n_{2.k}n_{.1k}n_{.2k}}{n_{..k}^2(n_{..k}-1)} = \frac{(20*5*5*20)}{25^2(25-1)} = 0.667
\]

Similarly, we compute the intermediate results for strata ($B=1$, $C=0$) and ($B=1$, $C=1$), and obtain
$\MH(A, Y)=17.5 > 3.84$. So $A$ and $Y$ have a causal relationship based on the test (for ${\alpha}=0.05$ or $\chi^2_{\alpha}=3.84$).

In the context $A=0$, we test if $B$ and $Y$ have a causal relationship, based on the following summaries of the data set:
\vspace{0.1cm}
\begin{center}
\begin{tabular}{|c|cc|}
\hline
\scriptsize $C$ & \multicolumn{2}{|c|}{\scriptsize Y} \\[-3pt]
 \scriptsize $0$ & \scriptsize$1$ &\scriptsize $0$ \\[-2pt]
\hline
\scriptsize $B=1$ &\scriptsize 0 &\scriptsize 10 \\[-2pt]
\scriptsize $B=0$ &\scriptsize 10 &\scriptsize 0 \\
\hline
\end{tabular}
\hspace*{0.2 cm}
\begin{tabular}{|c|cc|}
\hline
\scriptsize $C$ & \multicolumn{2}{|c|}{\scriptsize Y} \\[-3pt]
\scriptsize $1$ & \scriptsize$1$ &\scriptsize $0$ \\[-2pt]
\hline
\scriptsize $B=1$ &\scriptsize 0 &\scriptsize 20 \\[-2pt]
\scriptsize $B=0$ & \scriptsize 20 &\scriptsize 0 \\
\hline
\end{tabular}
\end{center}

\vspace{0.1cm}
From the above tables, we have $\MH(B,\!Y)\!=\!49\!>\!3.84$ in the context specific data set for $A\!=\!0$. So we can conclude that $B$ and $Y$ have a causal relationship in the context of $A\!=\!0$.
\end{example}

\section{Causal decision tree algorithm}
Normal decision trees have the following advantages: (1) The divide and conquer strategy of decision tree induction is very efficient. A decision tree construction algorithm is scalable to the data set size and the number of attributes. This is a major advantage in exploring big data; (2) Decision trees explore both global and context specific relationships, and the latter provide refined explanations for the former. They jointly provide comprehensive explanations for a data set.

Therefore in this paper we exploit these advantages for exploring causal relationships. However, the challenges for building a CDT include: (1) The criterion for choosing a branching attribute for a normal decision tree needs to be replaced by a causality based criterion. In our algorithm, we make use of the Mantel-Haenszel test, a statistically sound method for testing causal signals; (2) The time complexity for Mantel-Haenszel test is quadratic to the size of a data set since all strata must be found in the first place. We use quick sort to facilitate the discovery of strata, which reduces the time complexity greatly.

Based on the above discussion, we present the CDT construction algorithm as shown in Algorithm~\ref{alg_CDT1}.

The algorithm takes 3 inputs: the data set $\textbf{D}$ for a set of predictive attributes $\textbf{X}$ and one outcome attribute $Y$; user specified confidence level for Mantel-Haenszel test and correlation test (to find relevant attributes); and the maximum height of the CDT. Having a maximum tree height makes the tree more interpretable. If we do not restrict the tree height, we can get a context which includes many attributes, and a causal relationship in such a context only explains a very specific scenario and has less interest to users. In our experiments, however, the height of a tree is short after pruning even we set the maximum height of a tree high. So in practice it is not necessary to set the maximum height high.

Algorithm~\ref{alg_CDT1} firstly initiates the CDT (i.e. $\textbf{T}$), and sets the count of the height of the tree (i.e. $h$) as zero in Line 1. Then the functions TreeConstruct and TreePruning are called subsequently. Finally, the CDT is returned.

The treeConstruct function uses a recursive procedure to construct a CDT and it takes 5 inputs: current node $N$ to be expanded or terminated; the set of attributes $\textbf{Z} \subseteq \textbf{X}$ to expand the current subtree (whose root is $N$) and $\textbf{Z}$ contains only the attributes that have not been used in the tree; the context specific data set $\textbf{D}'$, where the context is the value assignments along the path from the root to $N$ (inclusive); $h$, the current height of the tree up to $N$; and $e$, the label of the edge from $N$ to the next node to be expanded.

Lines 1 to 4 of the treeConstruct function terminate $N$ if there is no attribute left in $\textbf{Z}$ and/or the depth of $N$ reaches the maximum height of the tree. $N$ is terminated by attaching to it a pair of leaves with edges of 1 and 0 respectively and labelling the leaves with the most probable values in $(\textbf{D}'|1)$ and $(\textbf{D}'|0)$ respectively.

If $N$ is not to be terminated, Line 5 finds a set of attributes correlated with $Y$ in the current context specific data set $\textbf{D}'$. This correlated attribute set is used to stratify $\textbf{D}'$ for the Mantel-Haenszel test. The reason for choosing a set of correlated attributes for stratification is discussed in Section~\ref{sec_falseDiscovery}.

In Lines 6 to 8, the partial association between $Y$ and each attribute in $\textbf{Z}$ is tested. The attribute, $W$ that has the most significant partial association with $Y$ (i.e. has the largest Mantel-Haenszel test statistic) is selected in Line 9. If the partial association between $W$ and $Y$ is insignificant, in Lines 10 to 13 we terminate $N$ by attaching a pair of leaves with edges of 1 and 0 respectively and labelling the leaves with the most probable values in data sets $(\textbf{D}'|1)$ and $(\textbf{D}'|0)$ correspondingly. If the partial association is significant, $W$ is a context specific cause of $Y$ and $W$ is added to the tree in one of the following two ways. If $e=null$, $W$ is set as the root of tree $\textbf{T}$; otherwise, $W$ is added as a child node of $N$ and the edge between $N$ and $W$ is labelled as $e$. Line 19 removes $W$ from $\textbf{Z}$ so it will not be used in the subtree again. In Lines 20 to 22, TreeConstruct is called recursively for $W$ with the context specific data sets $(\textbf{D}'|W=w)$ where $w \in \{0, 1\}$.

\begin{algorithm}[t!]
\begin{footnotesize}
\caption{\textbf{C}ausal \textbf{D}ecision \textbf{T}ree (CDT)}\label{alg_CDT1}
Input: $\textbf{D}$, a data set for the set of predictor attributes $\textbf{X}=\{X_1, X_2, \ldots, X_m\}$ and the outcome attribute $Y$, $h_{\max}$ the maximum height the tree, and $\alpha$, significance level for the Mantel-Haenszel (partial association) test and correlation test.\\
Output: $\textbf{T}$, causal decision tree
\end{footnotesize}
\begin{algorithmic}[1]
\begin{footnotesize}
\STATE {let $\textbf{T}=\emptyset$ and $h=0$}
\STATE {TreeConstruct($T$, $\textbf{X}$, $\textbf{D}$, $h$, null)} $\slash \slash$  $T$ is the root of $\textbf{T}$
\STATE {TreePruning($\textbf{T}$)}
\STATE return $\textbf{T}$
\end{footnotesize}
\end{algorithmic}
\begin{footnotesize}
\textbf{TreeConstruct}($N$, $\textbf{Z}$, $\textbf{D}'$, $h$, $e$)
\end{footnotesize}
\begin{algorithmic}[1]
\begin{footnotesize}
\IF {$\textbf{Z} = \emptyset$ OR $++h = h_{\max}$}
\STATE add two leaf nodes to $N$ with edges $e \in \{1, 0\}$ and label each with the most probable value of $Y$ in $(\textbf{D}'|N=e)$
\STATE return
\ENDIF
\STATE find a set of correlated attributes in $\textbf{Z}$ with $Y$ in $\textbf{D}'$
\FOR {each correlated attribute $X_i$}
\STATE compute $\MH(X_i, Y)$ in $\textbf{D}'$ stratified by the remaining correlated attributes
\ENDFOR
\STATE find attribute $W$ with the highest partial association
\IF {partial association between $W$ with $Y$ is insignificant}
\STATE add two leaf nodes to $N$ with edges $e \in \{1, 0\}$ and label each with the most probable value of $Y$ in $(\textbf{D}'|N=e)$
\STATE return
\ENDIF
\IF{$e = null$}
\STATE let node $W$ be the root of $\textbf{T}$
\ELSE
\STATE add node $W$ as a child node of $N$ and label the edge between $N$ and $W$ as $e$
\ENDIF
\STATE remove $W$ from $\textbf{Z}$
\FOR {each $w \in \{0, 1\}$}
\STATE call TreeConstruct($W$, $\textbf{Z}$, $(\textbf{D}' | W=w)$, $h$, $w$)
\ENDFOR
\end{footnotesize}
\end{algorithmic}
\begin{footnotesize}
\textbf{TreePruning}($\textbf{T}$)
\end{footnotesize}
\begin{algorithmic}[1]
\begin{footnotesize}
\FOR {each leaf in $\textbf{T}$}
\IF {its sibling leaf has the same label of $Y$ value as its}
\STATE change their parent node as a leaf node and label the leaf node with the common label
\STATE remove both leaves
\ENDIF
\ENDFOR
\end{footnotesize}
\end{algorithmic}
\end{algorithm}

The TreePruning function prunes leaves that do not have distinct labels. The function back traces the tree from the leaf nodes. When two sibling leaves of a parent node share the same label, their parent is converted to a leaf node and is labelled with the same label as their children in Line 3. Both leaves are then pruned in Line 4.

The time complexity of Algorithm \ref{alg_CDT1} mainly attributes to 3 factors: tree construction, forming strata, and causal tests.

For tree construction, at each split, firstly, we test the correlation of each (unused) attribute with $Y$, and the complexity is $O(mn)$  where $m$ is the number of predictive attributes and $n$ is the number of samples in the given data set. Then Mantel-Haenszel tests with $Y$ are conducted for all (relevant) attributes, and this is the most expensive part of the algorithm. For each test, the context specific data set $\textbf{D}'$ is sorted and strata are found in the data set, which has a complexity of  $O(n\log n)$, and for all the tests at a split, the complexity is $O(mn\log n)$. At most we have $2^{h_{\max}}$ splits. When $h_{\max}$ is not big, it is a small number and let it be a constant $n_s$. Therefore the time complexity for tree construction is $O(mn\log n)$. For tree pruning, the algorithm traverses the tree once and merge the leaves with the same labels under a branching node, which takes a constant time (proportional to $n_s$).

Overall the time complexity for building a CDT is $O(mn\log n)$.

\section{Experiments}
To evaluate CDT, three sets of experiments are conducted.

Firstly we experiment with 2 real world and 1 synthetic data sets to show that CDT is able to identify more interpretable relationships when comparing to normal decision trees. The normal decision trees are built using the C4.5 algorithm~\cite{DataMiningBookHan2011} implemented in Weka~\cite{Weka}.  It is difficult to evaluate discovered causal relationships as for most real world data sets we do not have the ground truths (true causal relationships). It is also impossible to use a method for evaluating classifiers to assess causal discovery results, because a model containing no causal relationships may give accurate classification, such as the decision tree in Figure~\ref{fig_ex-decisionTree1}. Thus we take a common sense approach to do the evaluation by using two data sets from which the results could make sense to ordinary people. We examine the results to see if they are reasonable, and contrast the CDTs to normal decision trees built based on the data.

Secondly experiments with synthetic data sets are carried out to demonstrate the ability of CDT in finding causal relationships comparing to the commonly used Bayesian network learning method, the PC algorithm~\cite{Spirtes2001book}.

Finally we evaluate the scalability of CDT using synthetic data sets in comparison to C4.5 and PC.

In all the experiments, the significance level for Mantel-Haenszel tests is 0.05, and the maximum level of CDTs is 5.

\subsection{CDT finds more interpretable relationships}
\subsubsection{The Titanic data set}
This is a built-in data set in R (https://stat.ethz.ch/R-manual/R-devel/library/datasets/html/Titanic.html). It has ticket and gender information of passengers and crew on board Titanic. The outcome attribute is `survived' (or not). A summary of the data set (converted to a binary data set) is given in Table~\ref{tab_Titanic_Adult_data}. We try to establish relationships between `survived'  and the other attributes. 

\begin{table}
\center
\caption{Summary of Titanic and Adult data sets}
\label{tab_Titanic_Adult_data}
\begin{tabular}{|l|cc|c|}
\hline
\multicolumn{4}{|c|}{The Titanic data set}\\
\hline
Attributes & yes & no & comment \\
\hline
firstClass & 325 & 1876 & \\
secondClass & 285 & 1916 & \\
thirdClass & 706 & 1495 & \\
crew & 885 & 1316 & \\
female & 470 & 1736 & \\
survived & 711 & 1490 & outcome\\
\hline\hline
\multicolumn{4}{|c|}{The Adult data set}\\
\hline
Attributes & yes & no & comment \\
\hline
age $<$ 30 & 14515 & 34327 & young \\
age $>$ 60 & 3606 & 45236 & old \\
private & 33906 & 14936 & private company employer \\
self-emp & 5557 & 43285 & self employment \\
gov & 6549 & 42293 &  government employer \\
education-num$>$12 & 12110 & 36732 & education years\\
education-num$<$9 & 6408 & 42434 & education years\\
Prof & 23874 & 24968 & professional occupation \\
white & 41762 & 7080 & race \\
male & 32650 & 16192 & \\
hours $>$ 50 & 5435 & 43407 & weekly working hours \\
hours $<$ 30 & 6151 & 42691 & weekly working hours \\
US & 43832 & 5010 & nationality\\
$>$50K & 11687 & 37155 & annual income, outcome\\
\hline
\end{tabular}
\end{table}


\begin{figure}
\centering
\begin{tabular}{ll}
\hspace*{-0.5 cm}\includegraphics[width=0.24\textwidth]{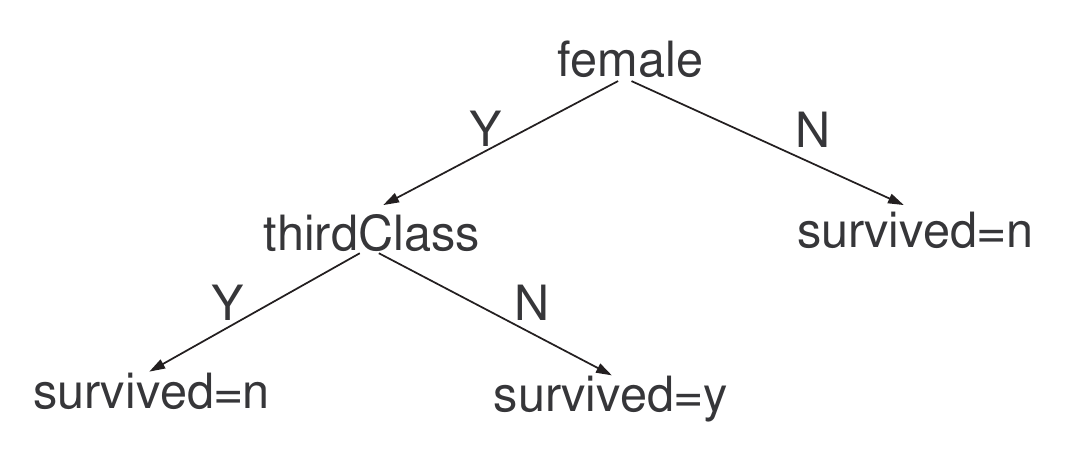} & \includegraphics[width=0.21\textwidth]{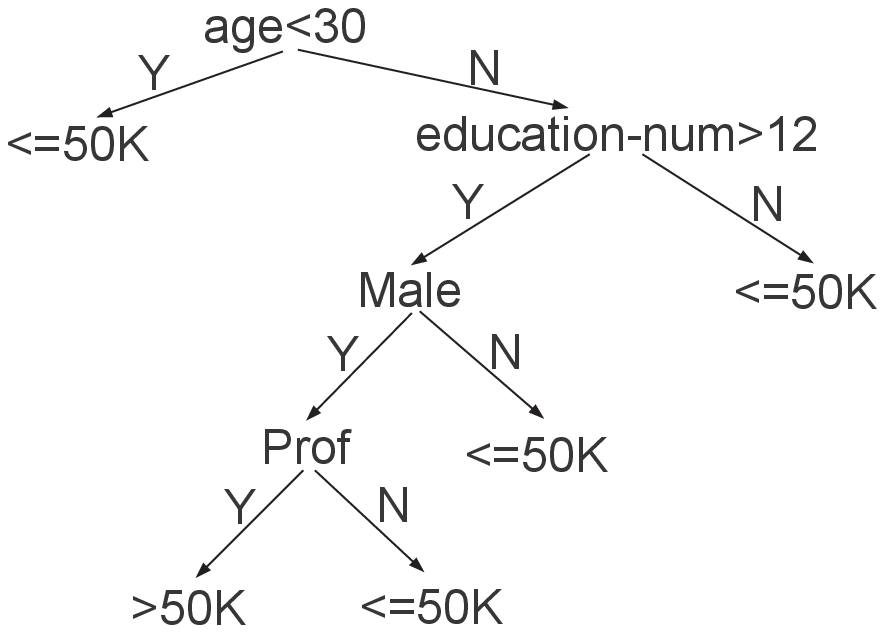} \\
\end{tabular}
\vspace{-0.15cm}
\caption{CDTs of the Titanic (left) and Adult (right) data sets}
\vspace{-0.2cm}
\label{fig_Titanic_Adult}
\end{figure}

The CDT built from the data set is shown in Figure~\ref{fig_Titanic_Adult} (left). At the first level, the tree reveals a causal relationship between `female' (gender) and `survived'. This relationship is sensible as we know that if someone was a female, she was likely to have higher priority to board the limited number of lifeboats. At the second level, the tree gives a context specific causal relationship between `thirdClass' and `survived' in the female group, which is reasonable too as passengers in the lowest class cabins would have less chance to escape. Therefore the tree is simple but it gives insights about the causes of surviving for people on Titanic, and the results are logic.



\subsubsection{Adult data set - census income}

The Adult data set (Table~\ref{tab_Titanic_Adult_data}) was retrieved from the UCI Machine Learning Repository \cite{Bache+Lichman:2013} and it is an extraction of 1994 USA census database. It is a well known classification data set to predict whether a person earns over 50K or not in a year. We recoded the data set to make the causes for high/low income more clearly and easily understandable. The objective is to find the causal factors of high (or low) income.



From the CDT built with the data set (Figure~\ref{fig_Titanic_Adult}, right), there is a causal relationship between `age$<$30' (or not) and income, i.e. young adults have lower income, which follows common knowledge. For older adults, year of education is causally related to income, i.e. adults with education shorter than 12 years would have low income, which makes good sense too. For older and highly educated adults, gender affects income such that females have lower income than male (an unfortunate finding but it could be true in reality). In the highly educated and older male group, occupation is a causal factor of income so that those in professional occupations earn more than those not in professional occupations. We see that the CDT gives sensible explanations for the causes of high or low income.

\begin{figure}
\center
      \includegraphics[width=0.45\textwidth]{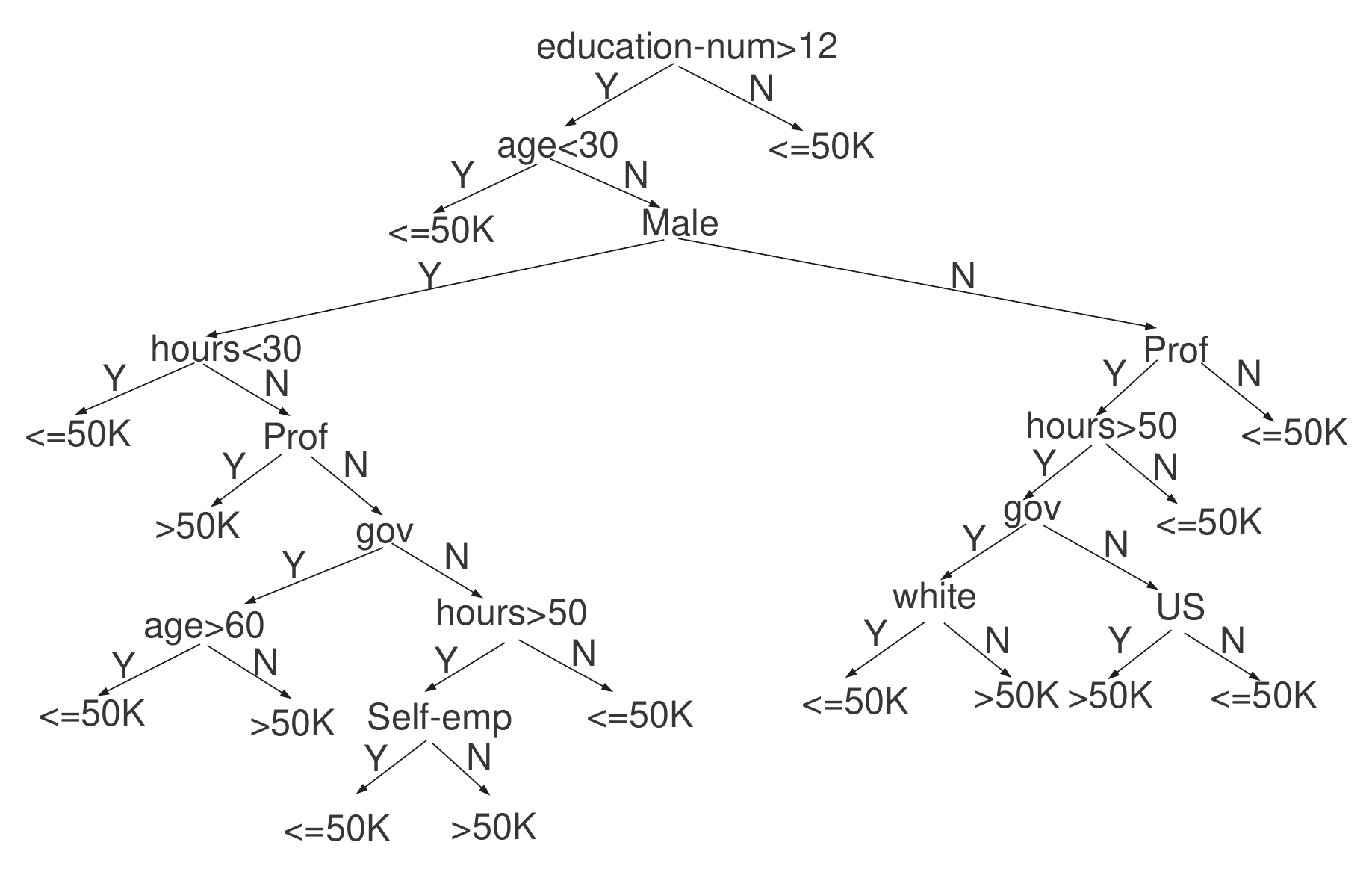}
\vspace{-0.15cm}
\caption{A decision tree of Adult data set}
\vspace{-0.2cm}
\label{fig_AdultC45}
\end{figure}

From the normal decision tree from the Adult data set (Figure~\ref{fig_AdultC45}), we have observed that: (1) A normal decision may be large for high classification accuracy but a large tree reduces its interoperability. The objectives of causal discovery and classification may not be consistent;
(2) Causality based and classification based criteria do not make the same choice. In the top level, the branching attribute of the normal decision tree is `education-num$>$12' while the branching attribute of the CDT is `age$<$30'. From common knowledge, age should have stronger influence on income than years of education since young adults usually get lower income in most cases, simply because of their lack of working experience. 
Formally, there are more strata (13.3\% of all strata) violating the causal relationship between `education-num$>$12' and `$>$50K' than those (7.75\% of all strata) violating the causal relationship between `age $<$30' and `$<$ 50K'.
This is why the CDT chooses `age $<$30' as the root. In contrast, since attribute `education-num $>$ 12' has a higher information gain than `age $<$30' it is chosen to split the data set firstly in a normal decision tree. Different choices lead to different trees. In this example, the different choices do not cause significant difference as `age $<$30' is chosen immediately after `education-num $>$ 12', but in other data sets, the difference could be significant. For causal discoveries, it is better to choose a causal based criterion.

\begin{figure}
\center
      \includegraphics[width=0.45\textwidth]{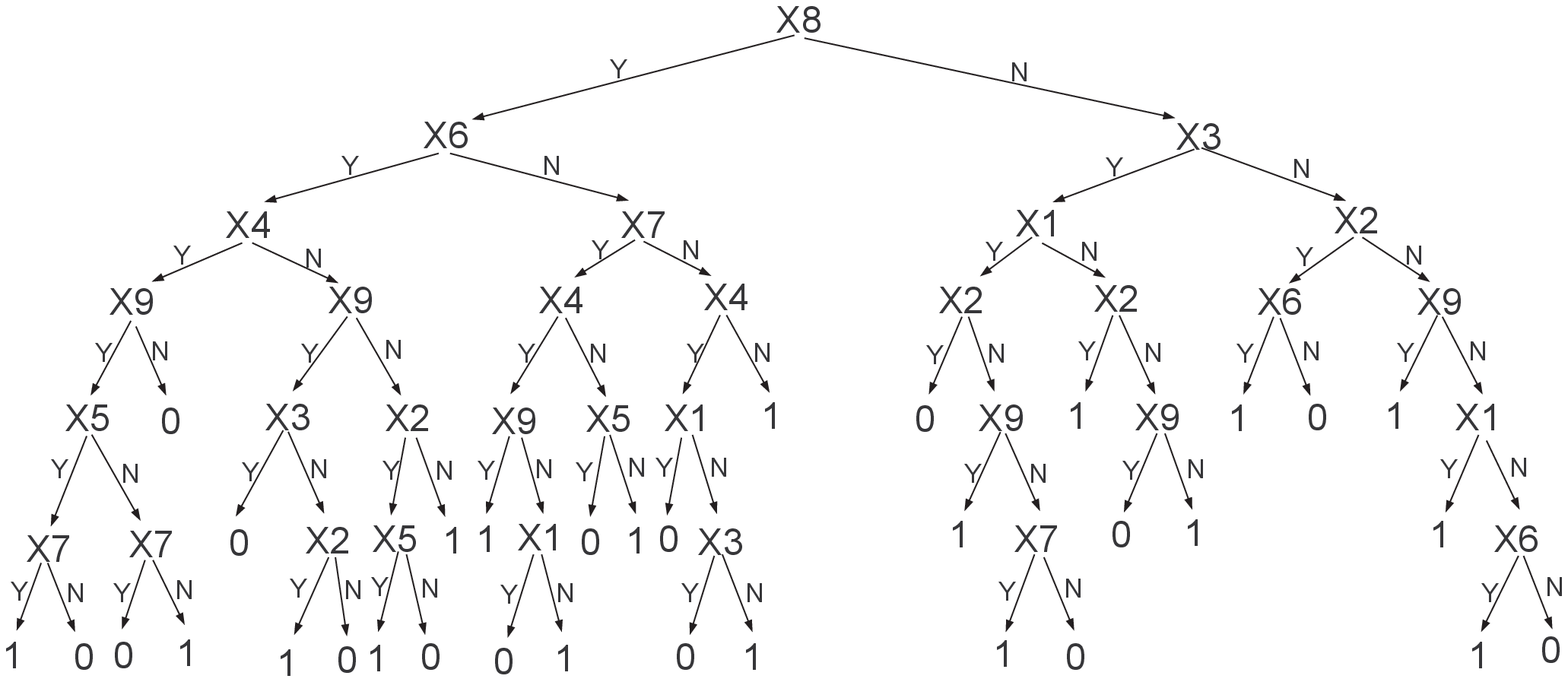}
\vspace{-0.15cm}
\caption{A decision tree of a randomly generated data set}
\label{fig_RandomDataC45}
\vspace{-0.2cm}
\end{figure}

With the above data sets, the first few levels at the top of a normal decision tree is quite interpretable since the causal relationships are evident in data. A CDT and a normal decision tree will be different when the causal relationships are subtle or with noises. To demonstrate this point, we build a CDT and a normal decision tree with a randomised data set where there is no relationship at all. Values in each of 10 attributes were randomly drawn with 50\% 1s and 50\% 0s in the data set. When we tried to learn a CDT from the data, no tree was returned and this is expected. However, C4.5 grew a decision tree as in Figure~\ref{fig_RandomDataC45}.

This result shows that the relationships in a normal decision tree may not be meaningful at all and a more interpretable decision tree, like a CDT, is necessary.

\subsection{CDT identifies causal relationships}\label{exp_CDT_causal}
\subsubsection{Finding global causal relationships}
To show that CDT is competent to discover causal relationships, we use 5 groups of synthetic data sets, each group containing 10 data sets with the same number of variables, to compare the findings of CDT and the PC algorithm from the data. In total 50 data sets are used, and each data set contains 10k samples.

The data sets are generated using the TETRAD tool (http://www.phil.cmu.edu/tetrad/). To create a data set, in TETRAD we firstly generate randomly a causal Bayesian network structure with the specified number of variables (20, 40, 60, 80, or 100), and randomly select a node with a specified degree (i.e. number of parent and children nodes, which is in the range of 3 to 7) as the outcome attribute for the data set. The conditional probability tables of the causal Bayesian network are also randomly assigned. The data set is then generated using the built-in Bayes Instantiated Model (Bayes IM) based on the conditional probability tables. The ground truth of the data is the set of nodes directly connected to the outcome variable in the causal Bayesian network structure.   

We then apply CDT and PC to each of the 50 data sets, and for each group of the data sets, the average recalls of the algorithms are shown in Table \ref{tableAvgRecall} (Part A). It can be seen that in general CDT can detect similar percentages of causal relationships as PC does, indicating that CDT has comparable ability and has obtained consistent results in discovering causal relationships as the commonly used approach. We are aware that the causal relationships identified by CDT are context specific while those discovered by PC is global or context free. However, it is reasonable to assume that if a causal relationship exists with no context, it should appear in the contexts too, and these relationships have been mostly picked up by the CDTs. 

\vspace{-0.1cm}
	\begin{table}[h]
		\caption{Average recalls of CDT and PC} 
		\centering 
		\begin{tabular}{c c c c c c} 
		\hline
		\multicolumn{6}{c}{$\textbf{Part A}$: Average recall of global causal relationships}\\
			\hline\hline 
			Group & \#D & \#V & Recall (CDT) & Recall (PC)\\ [0ex] 
			\hline 
			1 & 10 & 20 & 85\% & 75\%\\
		2 & 10 & 40 & 77\% & 79\%\\
			3 & 10 & 60 &  89\% & 78\%\\
				4 & 10 & 80 & 94\% & 90\%\\
		5 & 10 & 100 & 85\% & 94\%\\\hline
			\multicolumn{6}{c}{$\textbf{Part B}$: Average recall of context specific causal relationships}\\
			\hline\hline
					Group & \#D & \#V & Recall (CDT) & Recall (PC)\\ [0ex] 
			\hline 
			6 & 10 & 20 & 81\% & n/a\\
		7 & 10 & 40 &  77\% & n/a\\
			8 & 10 & 60 & 85\% & n/a\\
				9 & 10 & 80 & 89\% & n/a\\
		10 & 10 & 100 & 78\% & n/a\\\hline
			\multicolumn{6}{l}{\footnotesize{\#D:number of data sets in a group}}\\
			\multicolumn{6}{l}{\footnotesize{\#V: number of variables in one data set}} \\
		   \hline
		\end{tabular}
		\label{tableAvgRecall} 
	\end{table}

\subsubsection{Finding context specific causal relationships}
In order to test the performance of CDT in finding context specific causal relationships, we also use 5 groups of synthetic data sets, each group containing 10 data sets with the same number of variables (20, 40, 60, 80 or 100).

To create a data set, e.g. with 20 binary variables, $\{v_1, v_2, \ldots, v_{20}\}$, we firstly create a causal Bayesian network structure that contains only one edge, e.g. between $v_1$ and $v_{20}$, and all other nodes are isolated nodes. Based on this structure, we use logistic regression to simulate the data set for the Bayesian network. One of the two causally related variables, e.g. $v_{20}$ is chosen as the outcome variable, then $v_1$ in this example is the ground truth of the global causal node of $v_{20}$. However, we do not know the ground truth of the context specific causal relationships around $v_{20}$. Our solution is to use $v_1$ as the context variable, and apply PC-select \cite{pcdag} (also known as PC-simple \cite{Buehlmann2010}) to the two partitions of the data set respectively, one partition containing all the samples with ($v_1=0$) and one containing all the samples with ($v_1=1$) (while the $v_1$ column is excluded). In this way, we identify the variables that are causally related to $v_{20}$ within each of the two contexts, ($v_1=0$) and ($v_1=1$), and use the findings as the ground truth of the context specific causal relationships around $v_{20}$ in the data set. PC-select is a simplified version of the PC algorithm for finding causal relationships around a given outcome variable.


We then apply CDT to each of the 50 data sets generated. The CDT built from each of the data sets always has the node that is causally related to the output attribute as its root, i.e. the CDT correctly finds the global causal relationship. Moreover, each of the CDTs also contains context specific causal relationships.
We did not prune CDT trees in these data sets since some randomly generated data sets have skewed distribution, which makes the pruning too aggressive. We will design a  pruning strategy for skewed data sets in future work.
Table \ref{tableAvgRecall} (Part B) summarises the average recall of CDT in finding the context specific causal relationships. From the table, CDT is able to discover the majority of the context specific causal relationships. PC, in contrast, does not find any context specific causal relationships in the data sets since it is not design for the purpose. If we want to use PC to find the context specific causal relationships, we have to run PC in each context specific data set, which is impractical. On the other hand, CDT can find context specific causal relationships in the complete data sets.

\subsection{Scalability of the CDT algorithm}

We test the scalability of the CDT algorithm by comparing it with the C4.5 \cite{DataMiningBookHan2011} algorithm implemented in Weka \cite{Weka} and the PC Algorithm~\cite{Spirtes2001book}. 

We use 12 synthetic data sets generated with the same procedure as for generating the data sets in Section \ref{exp_CDT_causal}-1). To be fair among data sets, we chose the nodes with the same degree as the target variables. The comparisons were carried out using the same desktop computer (Quad core CPU 3.4 GHz and 16 GB of memory).

\begin{figure}
\includegraphics[width=0.45\textwidth]{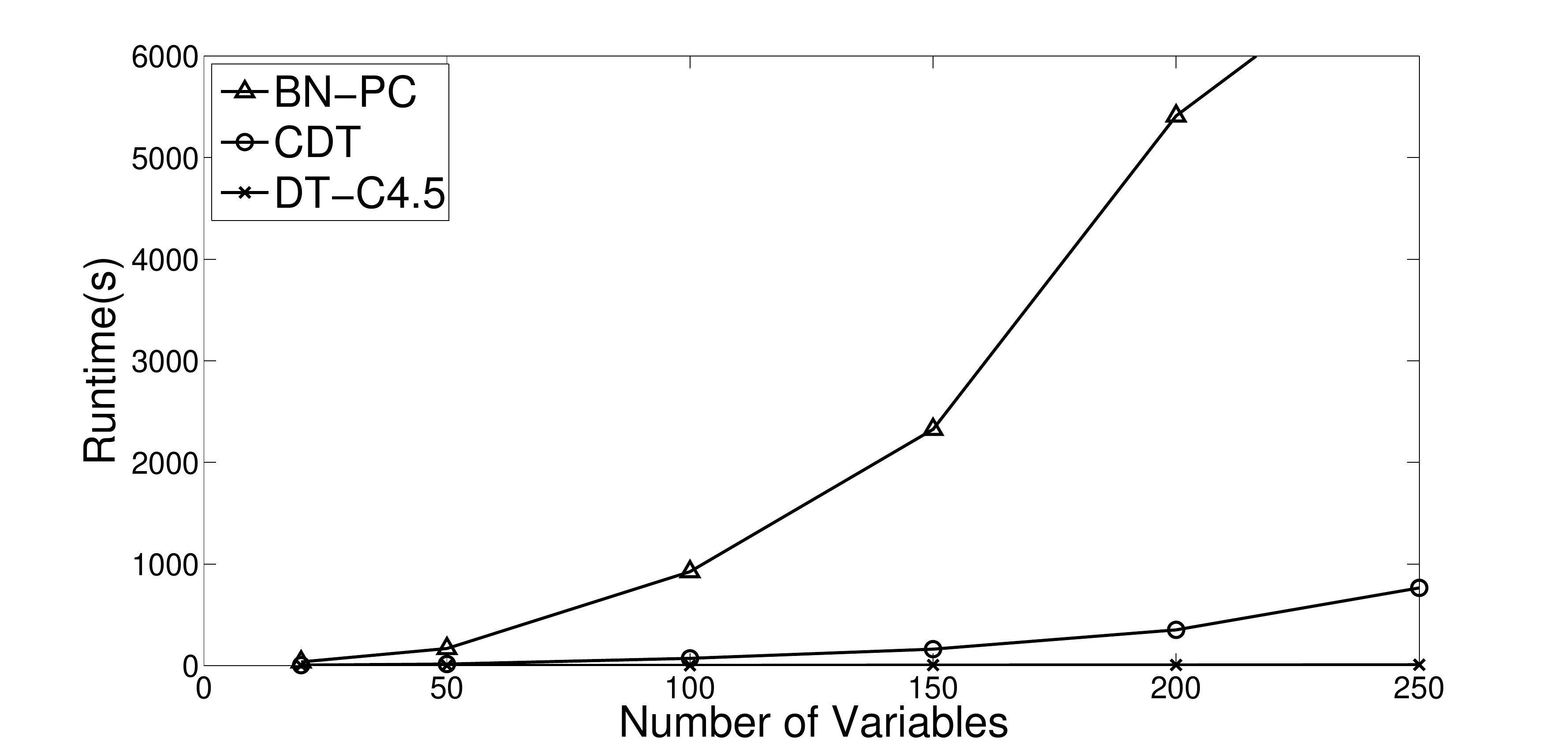}\\
\vspace{0.1cm}
\hspace{0.2cm}\includegraphics[width=0.4\textwidth]{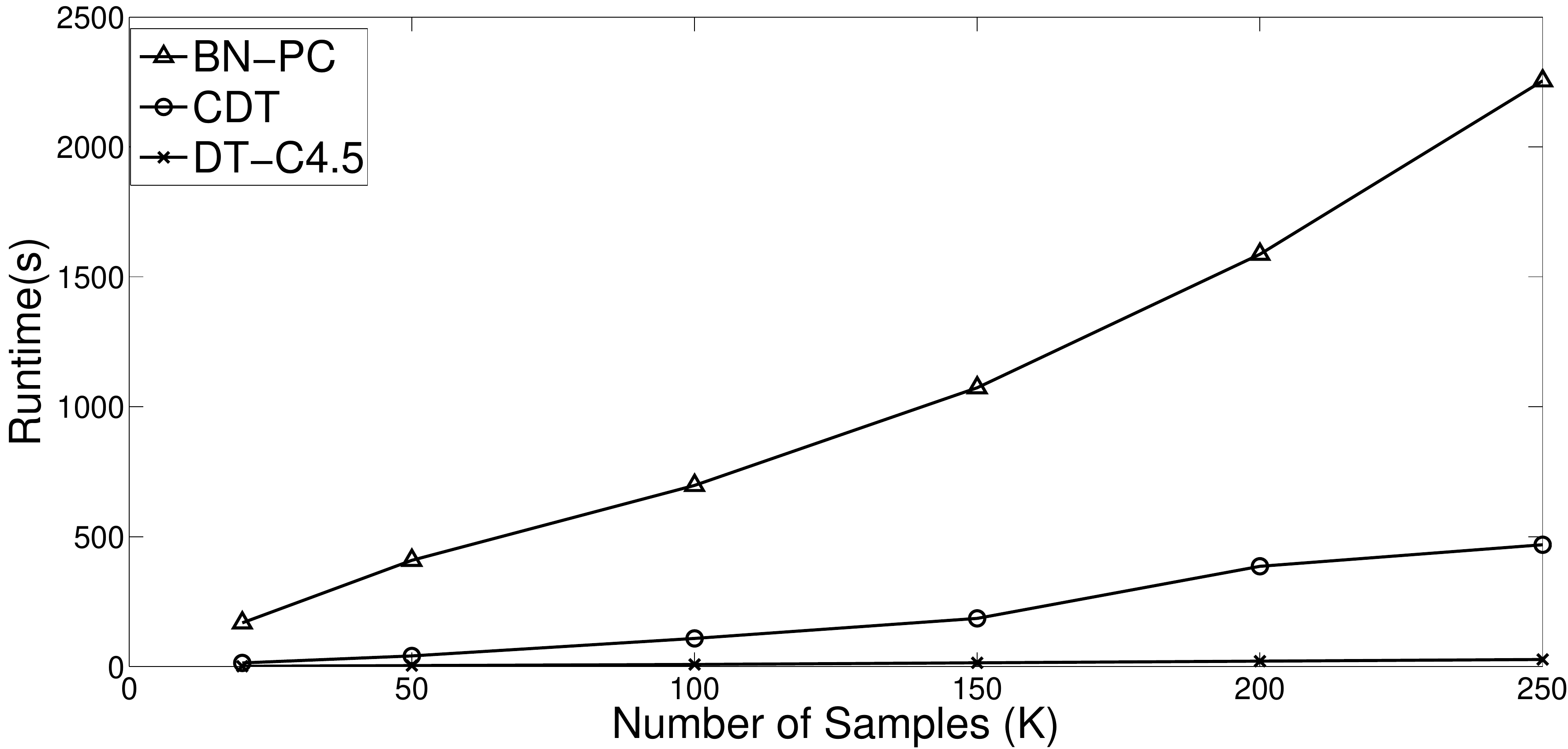}
\vspace{-0.15cm}
\caption{The scalability of CDT in comparison to C4.5 and PC}
\vspace{-0.2cm}
\label{fig_Scalability}
\end{figure}

The comparison results are shown in Figure~\ref{fig_Scalability}. The run time of CDT is almost linear to the size of the data sets and the number of attributes. It is less efficient than C4.5 but more efficient than PC. The results have shown that the proposed CDT is practical for high dimensional and large data sets.


\section{Discussions}
\subsection {Difference from other causal trees}\label{sec_otherCausalTree}

In this section, we will differentiate our CDTs from other causal trees derived from causal Bayesian networks, including the  conditional probability table tree (CPT-tree) \cite{Boutilier96context-specificindependence} and causal explanation tree \cite{Nielsen2008}.

A causal Bayesian network (CBN) \cite{Spirtes2010} consists of a causal structure of a directed acyclic graph (DAG), with nodes and arcs representing random variables and causal relationships between the variables respectively, and a joint probability distribution of the variables. Given the DAG of a CBN, the joint probability distribution can be represented by a set of conditional probabilities attached to the corresponding nodes (given their parents). A CBN provides a graphical visualisation of causal relationships, a reasoning machinery for deriving new knowledge (effects) when evidence (changes of causes) is fed into the given network; as well as a mechanism for learning causal relationships in observational data. In recent decades, CBNs have emerged, especially in the areas of machine learning and data mining, as a core methodology for causal discovery and inference in data.

\begin{figure}
\center
      \includegraphics[width=0.35\textwidth]{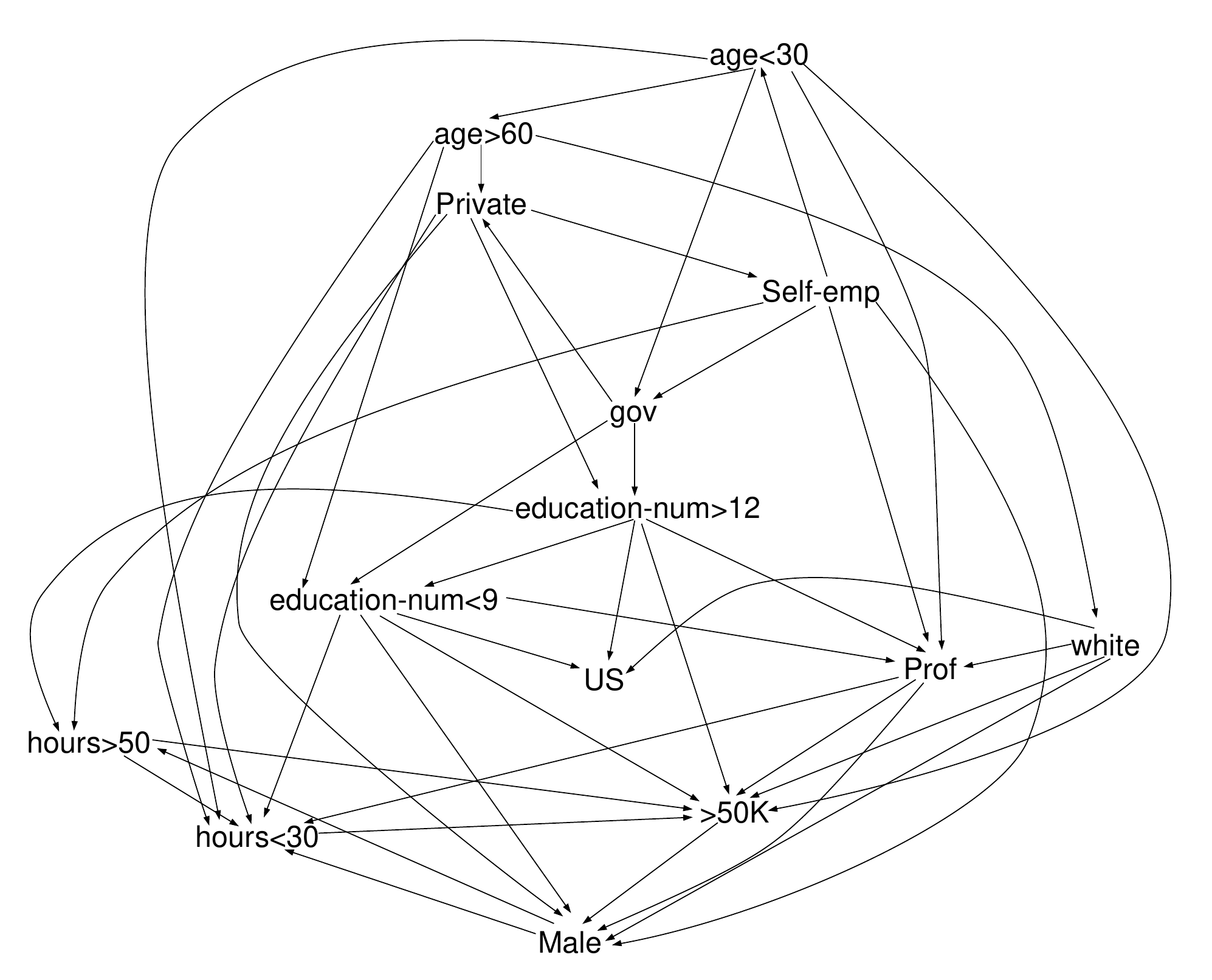}
\vspace{-0.15cm}
\caption{A causal Bayesian network of the Adult data set}
\vspace{-0.2cm}
\label{fig_AdultBayesinnetwork}
\end{figure}

A CBN depicts the relationships of all attributes under consideration, and it can be complex when the number of attributes is more than just a few. For example, it takes some effort to understand the CBN in Figure~\ref{fig_AdultBayesinnetwork} learnt from the Adult data set, even though there are only 14 attributes in the data set. A CBN does not give a simple model to explain the causes of an outcome as our CDT does.

\begin{figure}
\center
      \includegraphics[width=0.45\textwidth]{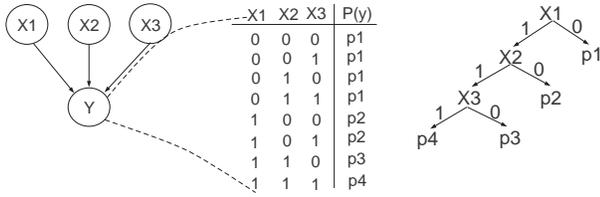}
\vspace{-0.15cm}
\caption{An illustration of CPT-tree. (L): A Bayesian network; (M): Conditional probability table of $Y$; (R): CPT-tree.}
\vspace{-0.5cm}
\label{fig_CPTTree}
\end{figure}

The conditional probability table tree (CPT-tree) \cite{Boutilier96context-specificindependence} is designed to summarise the conditional probability tables of a CBN for concise presentation and fast inference. An example of CPT-trees is shown in Figure~\ref{fig_CPTTree}. The probabilistic dependence relationships among the outcome $Y$ and its parent nodes $X_1, X_2$ and $X_3$ (causes of $Y$) are specified by a conditional probability table where the probabilities of $Y$ given all value assignments of its parents are listed. The size of a conditional probability table is exponential to the number of parent nodes of $Y$ and can be very large. For example, for 20 parent nodes, the conditional probability table will have 1,048,576 rows. This table will be difficult to display and the inference based on the table is inefficient too. Given a context, i.e. one or more parent nodes taking an assignment of a value, the probability of $Y$ may be constant (without being affected by the values of other parents). So a conditional probability table can be represented clearly with a tree structure, called a conditional probability table tree (CPT-tree), as illustrated in Figure \ref{fig_CPTTree}. In the CPT-tree, the causal semantics is naturally linked to the CBN where all parent nodes are direct causes of variable $Y$.

There are two major differences between a CPT-tree and a CDT. Firstly, CPT-trees are built from CBNs and CDTs are built from data sets directly. Before building the CPT-trees, we already know the causal relationships, and a CPT tree specifies how the assignments of some causal variables link to outcome values. This is impractical in many real world applications since we do not know the CBN or we could not build a CBN from a data set, particularly a large data set, as existing algorithms for learning CBNs cannot handle a large number of variables and they often only present a partially oriented CBN. Secondly, in a CBN, the parents of a node $Y$ are all global causes of $Y$. As a CPT-tree is derived from a CBN, all the variables included in a CPT-tree are all global causes. However, as discussed previously, it is possible that under a context, a variable becomes causally related to $Y$. For example, in the Titanic data set, `thirdClass' cabin is not a causal factor in the whole data set (i.e. it is not a global cause of `survived'), but it becomes a causal factor in the context of female passengers/crew. So such causal relationships will not be discovered or represented by a CBN and thus not by the CPT-trees too, but they can be be revealed and represented by our CDTs.


A causal explanation tree \cite{Nielsen2008} aims at explaining the outcome values using a series of value  assignments of a subset of attributes in a CBN. A series of value assignments of attributes form a path of a causal explanation tree, and a path is determined by a causal information flow. The assignment of a set of attributes along a path represents an intervention in the causal inference in a CBN. The causal interpretation is based on the causal information flow criterion used for building a causal explanation tree. However this method is impractical since we do not have a CBN in most real world applications as explained previously. Similarly a causal explanation tree cannot capture the context specific causal relationships encoded in a CDT, because the explanation tree is obtained from a CBN, which only encodes global causes.

\subsection {Assumptions and practical considerations}
\label{sec_falseDiscovery}

Causal discovery is based on assumptions. In the causal Bayesian network discovery framework, some assumptions, such as causal Markov condition, faithfulness and causal sufficiency~\cite{Spirtes2001book}, are used to ensure the causal semantics of the discoveries. Simply speaking, Markov condition requires that every edge in a causal Bayesian network implies a probabilistic dependence. The faithfulness assumption ensures that for two variables that are probabilistically dependent, there is a corresponding edge between the two variables in the causal Bayesian network. The causal sufficiency assumes that there is no unmeasured or hidden causes in data. Up to now, we have not explicitly discussed the causal assumptions. However, we do need certain assumptions which will be discussed in the following.

The causal interpretation of a CDT is ensured by the evaluation in the stratified data sets of the difference in the potential outcomes of a possible causal attribute $X_i$. In each stratum, the individuals are indistinguishable, or the attributes possibly affecting the outcome $Y$ take the same values and they do not affect the estimation of the causal effect of $X_i$ on $Y$. Therefore, the causal effect estimated using the stratified data sets approaches the true causal effect.

An assumption here is that the differences of individuals should be captured by the set of attributes used for stratification.
This assumption implies causal sufficiency that all causes are measured and included in the data set.
A na{\"{i}}ve choice is to select all attributes other than the attribute being tested ($X_i$) and the outcome ($Y$), for stratification. However, this is not workable for high dimensional data sets since many strata will contain very few or no samples when the number of attributes is large. As a result, the CDT algorithm may not find any causal relationship. For example, diverse information, such as demographic information, education, hobbies and liked movies, is collected as personal profile in a data set. However, if all the attributes are used for stratification, they reduce the chance of finding sizable strata for reliable discovery. In fact, it is unwise to use any irrelevant attributes, such as hobbies and liked movies, for stratification when the objective is to study, e.g. the causal effect of a treatment on a disease.

A reasonable and practical choice of stratifying attributes is the set of attributes that may affect the outcome, called relevant attributes in this paper. Differences in irrelevant attributes that do not affect the outcome should not impact the estimation of the causal effect of the studied attribute on the outcome. For example, different hobbies and liked movies may not affect the estimation of the causal effect of a treatment on a disease. Therefore, only those relevant attributes should be used for stratifying a data set, and this is what we have used in the algorithm for building a CDT. In case there are many relevant variables, which may result in many small strata, we restrict the maximum number of relevant variables to ten according the strength of correlation. The purpose of this work is to design a fast algorithm to find causal signals in data automatically without user interactions. We do tolerate certain false positives and expect that a real causal relationship will be refined by a dedicated follow-up observational study.

In many real world studies, the stratification is based on a limited number of demographic attributes, e.g. gender, age group and residential areas. Thinking about a heath study, it is very difficult to recruit volunteers with the same background (age, diet, education, etc.), and stratification on more than a few attributes is just impractical. Considering some stratifying attributes is better than considering none.

CDTs help practitioners with the discovery of causal relationships in the following ways although it may not confirm causal relationships: (1) Because of stratification, many spurious relationships that are definitely not causal will be excluded from the resulting CDTs, so practitioners will have a smaller set of quality hypotheses for further studying; (2) Context specific causal relationships are more difficult to be observed than global causal relationships. CDTs are useful for practitioners to find hidden context specific causal hypotheses.

\section{Related work}

Discovering causal relationships in passively observed data has attracted enormous research efforts in the past decades, due to the high cost, low efficiency and unknown feasibility of experiments based approaches, as well as the increasing availability of observational data. To the credit of the theoretical development by a group of statisticians, philosophers and computer scientists, including Pearl \cite{Pearl2000}, Spirtes, Glymour \cite{Spirtes2001book} and others, we have seen graphical causal models playing dominant role in causality discovery. Among these graphical models, causal Bayesian networks (CBNs) \cite{Spirtes2010} have been the most developed and used one.


Many algorithms have been developed for learning CBNs \cite{Neapolitan2003,Spirtes2010}. However in general learning a complete CBN is NP-hard \cite{Chickering2004} and the methods are able to handle a CBN with only tens of variables, or hundreds if the causal relationships are sparse \cite{Spirtes2010}.

Consequently, local causal relationship discovery around a given target (outcome) variable has been actively explored recently as in practice we are often more interested in knowing the direct causes or effects of a variable, especially in the early stage of investigations. The work presented in this paper is along the line of local causal discovery.

Existing methods for local causal discovery around a given a target fall into two broad categories: (1) Methods that adapt the algorithms or ideas for learning a complete CBN into local causal discovery, such as PC-Simple \cite{Buehlmann2010,pcdag}, a simplified version of the well-known PC algorithm \cite{Spirtes2001book} for CBN learning; and HITON-PC \cite{Aliferis2010}, which applies the basic idea of PC to find variables strongly (and causally) related to a given target; (2) Methods that are designed to exploit the high efficiency of popular data mining approaches and the causal discovery ability of traditional statistical methods, including the work in \cite{Jin2012} and \cite{LiCR2013}, both using association rule mining for identifying causal rules; and the decision tree based approach \cite{Frey2003} for finding the Markov blanket of a given variable.

The CDT proposed in this paper belongs to the second category, as it takes advantage of decision tree induction and partial association tests. Comparing to other methods in the category, however, the proposed CDT approach is distinct because it is aimed at finding a sequence of causal factors (variables along the path from the root to a leaf of a CDT) where a preceding factor is a context under which the following factors can have impact on the target, while the other methods identify a set of causal factors each being a cause or an effect of the given target, and they only discover global causal relationships. However, in practice, a variable may not be a cause of another variable globally, but under certain context, it may affect other variables. A CDT provides a way to identify such context specific causal relationships. Additionally because a context specific causal relationship contains information about the conditions in which a causal relationship holds, such relationships are more prescriptive and actionable and thus are more suitable for decision support and action planning.

In terms of using decision trees as a means for causality investigation, except from the above mentioned method for identifying Markov blankets \cite{Frey2003}, most existing work takes decision trees as a tool for causal relationship representation and/or inference, assuming that the causal relationships are known in advance. Examples include the CPT-trees \cite{Boutilier96context-specificindependence} and causal explanation tree \cite{Nielsen2008} introduced in the Discussions section, which are both derived from a known causal Bayesian network.
Unlike these trees, our CDT is mainly used as a tool for detecting causal relationships in data, without any assumption of known causal relationships.

\section{Conclusion}
In this paper, we have proposed causal decision trees (CDTs), a novel model for representing and discovering causal relationships in data.

A CDT provides a compact and precise graphical representation of the causal relationships between a set of predicate attributes and an outcome attribute. The context specific causal relationships represented by a CDT are of great practical use and they are not encoded by existing causal models.

The algorithm developed for constructing a CDT utilises the divide and conquer strategy for building a normal decision tree and thus is fast and scalable to large data sets. The criterion used for selecting branching attributes of a CDT is based on the well established potential outcome model and partial association tests, ensuring the causal semantics of the tree.

Given the increasing availability of big data, we believe that the proposed CDTs will be a promising tool for automated discovery of causal relationships in big data, thus to support better decision making and action planning in various areas.






%
\bibliographystyle{abbrv}


\end{document}